%% file: paper.tex
\documentclass[10pt, conference, compsocconf]{IEEEtran}

\usepackage[show]{chato-notes}
\usepackage{footnote}

\ifCLASSINFOpdf
   \usepackage[pdftex]{graphicx}
\else
\fi
\usepackage{amsmath}
\usepackage{array}

\usepackage{subcaption}
\usepackage{fixltx2e}
 \usepackage{dblfloatfix}

\usepackage{url}

\usepackage{algorithm}
\usepackage[noend]{algorithmic}
\usepackage{multirow}
\usepackage{paralist}
\usepackage{cite}
\usepackage{pifont}
\usepackage{flushend}
\usepackage{balance}
\usepackage{graphicx}
\usepackage{comment}

\newcommand{\argmin}{\arg\!\min}
\newcommand{\spara}[1]{\smallskip\noindent{\bf{#1}}}

\newtheorem{problem}{Problem}
\newtheorem{lemma}{Lemma}

\newcommand{\field}[1]{\ensuremath{\mathbb{#1}}}
\newcommand{\bfx}{\ensuremath{\mathbf{x}}}

\newcommand{\bfc}{\ensuremath{\mathbf{c}}}

\newcommand{\intra}{\operatorname{intra}}
\newcommand{\inter}{\operatorname{inter}}

\newcommand{\minDist}{\operatorname{minDist}}
\newcommand{\maxDist}{\operatorname{maxDist}}

\usepackage[numbers,sort&compress]{natbib}

\begin{document}






%


\title{A Robust Framework for Classifying Evolving Document Streams in an Expert-Machine-Crowd Setting}

%
%
%
%
%

%
\author{
\IEEEauthorblockN{Muhammad Imran}
\IEEEauthorblockA{Qatar Computing Research Institute, HBKU\\
Doha, Qatar\\
mimran@qf.org.qa}
\and
\IEEEauthorblockN{Sanjay Chawla}
\IEEEauthorblockA{Qatar Computing Research Institute, HBKU\\
Doha, Qatar\\
schawla@qf.org.qa}
\and
\IEEEauthorblockN{Carlos Castillo}
\IEEEauthorblockA{Eurecat\\
Barcelona, Spain\\
chato@acm.org}
}

\maketitle

\begin{abstract}
An emerging challenge in the online classification of social media data streams is to keep the categories used for classification up-to-date. In this paper, we propose an innovative framework based on an Expert-Machine-Crowd (EMC) triad to help categorize items by continuously identifying novel concepts in heterogeneous data streams often riddled with outliers.
%
%
We unify constrained clustering and outlier detection by formulating a novel optimization problem: COD-Means. We design an algorithm to solve the COD-Means problem and show that COD-Means will not only help detect novel categories but also seamlessly discover human annotation errors and improve the overall quality of the categorization process. Experiments on diverse real data sets demonstrate that our approach is both effective and efficient.

\end{abstract}


\IEEEpeerreviewmaketitle

\input{01-introduction}

\input{02-problemdef}
\input{03-algorithm}

\input{04-experiments}
\input{05-relatedwork}

\input{06-conclusions}

%

\footnotesize{
\bibliographystyle{IEEEtran}
\bibliography{references}
}

\end{document}

%% file: 01-introduction.tex

\section{Introduction}\label{sec:introduction}


The application that motivates our work is time-critical analysis of a social media stream. We consider a basic operation on this stream, which is to rapidly categorize messages into a series of classes of interest and also to capture novel emerging categories. 
This is a relevant problem during all sort of crises, such as mass convergence events and emergencies, including sudden-onset natural and man-made disasters. 
This problem, in practice, is addressed through automatic classification, crowdsourced classification, or a combination of both~\cite{imran_2013_engineering,quinn_2011_crowdflow}. 

We developed a system~\cite{imran2014aidr} that follows the latter approach.
The system combines human and machine intelligence to categorize crisis-related messages on Twitter during the sudden-onset of natural or man-made disasters. 
%
%
The system obtains labels (for the messages) from human workers (volunteers in this case) and trains machine learning classifiers. The trained classifiers then enable the automatic classification of subsequent incoming messages into the defined categories. We refer to this unique collaboration between domain experts, crowd volunteers and machine learning classification as the Expert-Machine-Crowd (EMC) framework. 

There are two important challenges that have emerged for such a system to work in an optimal
fashion. 

\spara{Defining the  categorization scheme:} While it is impossible to predict 
apriori all types of categories (e.g. previously unknown needs of affected people) that are likely to emerge during a crisis, there is increasing evidence that most disasters do have a lot in common~\cite{rodriguez_2006_handbook}. However, the dictionary of all possible information categories people write about in social media is potentially very large.
%
%
For instance, in an analysis in 2012 of social media during 4 disasters, 28 information categories were found in the messages posted on this platform~\cite{vieweg2012situational}. Having a large set of categories is problematic from the crowdsourcing point of view as unskilled annotators cannot distinguish between very fine-grained categories, which introduce labeling errors and also drastically limits the size of the annotator pool. A large number of categories also increases burn-in (performance reduction in the pool) and drop-out (annotators leaving the pool), particularly in volunteer crowdsourcing settings~\cite{imran2014aidr}.

\spara{Labeling Errors:} The dynamic nature, brevity and ambiguity of messages often lead to labeling errors by annotators which can drastically reduce the accuracy and thus usefulness of  the classifier. Experiments with real data have shown that even a few poorly labeled messsages
can often lead to dramatically divergent results.

We propose a new optimization problem {\tt COD-Means} (where COD stands for Constrained Outlier Detection) to address the problem of simultaneously discovering new categories and identifying labeling errors. Initial categories and labeling are expressed in terms of constraints and an outlier detection step is used for error detection. The outlier discovery through {\tt COD-Means} is carried out by extending the $k$-means{-}{-} algorithm~\cite{chawla2013k}. The cluster refinement process is carried out by using constrained clustering which allows the generation of new and semantically distinct categories. The expert then refines the categories based on the output of {\tt COD-means}. Later initializations of {\tt COD-means} (i.e. once the supervised learning system is trained) can be triggered by the expert after a fixed time-interval or on observing a low classification accuracy. In that case, the {\tt COD-Means} also uses machine classified items for which the machine confidence is high (e.g. $\geq$ 90\%).

We formally define the problem in the next Section~\ref{sec:problemdef}. 
A novel optimization formulation and the associated algorithm are the subject of Section~\ref{sec:algo}.
An extensive suite of experiments were carried out to test the EMC framework
and are described in Section~\ref{sec:experiments}. Related work is the subject of Section~\ref{sec:relatedwork} and we conclude in Section~\ref{sec:conclusions} with a summary.
%


%% file: 02-problemdef.tex

\section{Problem definition}\label{sec:problemdef}

We are given as input a data set $D$ of documents which have been categorized by crowdsourcing workers (not necessarily experts) or an automatic classifier into a taxonomy ${\cal T} = \{T_{1},\ldots,T_{k}, Z\}$  containing $|{\cal T}| = k + 1$ categories.
The taxonomy forms a partition of the documents: $A \cap B = \emptyset ~ \forall A, B \in {\cal T}, A \neq B$.

We call categories $T_i$ the pre-existing categories, which are defined by the expert before the data begins to arrive, based on background domain knowledge.

The category $Z$, instead, is the ``miscellaneous'' category, used for documents that do not fit in any of the $T_i$ categories. In practical cases for the domain in which we focus (disasters and social media), this category contains anywhere from about $10\%$ to $30\%$ of the messages~\cite{imran2015towards}.

Our task is to produce a new taxonomy
$${\cal T}' = \{ T'_1, \dots, T'_k, N_1, \dots, N_n, Z' \}$$
with the following characteristics:

\begin{itemize}
 \item There are $n$ new categories: $|{\cal T}'| = |{\cal T}| + n$.
 \item Pre-existing categories are only slightly modified: $T_i \approx T'_i ~~ \forall i = 1 \dots k$.
 \item New categories are different from previous pre-existing categories: $T_i \not\approx N_j ~~ \forall i = 1 \dots k, j = 1 \dots n$
 \item $|Z'| < |Z|$: the size miscellaneous category is reduced
\end{itemize}

At a high level, our algorithm attempts to partition $D$ into $k + n + 1$ categories, guiding the clustering process in such a way that $k$ of these categories resemble the original $T_i$, and none of the categories overlap with each other. 

%% file: 03-algorithm.tex

\section{Proposed solution}\label{sec:algo}
Our solution framework consists of three parts (i) the crowd effort and the supervised learning output (the populated taxonomy $\cal{T}$) are captured as must-link (ML) and cannot-link (CL) constraints~\cite{davidsonR05}; (ii) discovery of new categories is framed as a constrained clustering problem where the number of clusters specified is greater than the size of the original taxonomy $\cal{T}$; (iii) 
to identify discrepancies between human and machine annotation, a new clustering problem {\tt COD-Means} is defined. 
A $k$-means type  algorithm is proposed to solve {\tt COD-Means}.


\subsection{Constraints Formation}
Our objective is to infer new categories and we use constrained clustering
to encode the output of the crowd and the supervised learning (SL) process.
We form two types of constraints: Must-Link (ML) constraints
and Cannot-Link (CL) constraints. For the taxonomy $\cal{T}$ we formulate
these constraints as follows.
{\bf ML constraints:} data elements belonging to a category $T_{i}$ are encoded 
as ML constraints, i.e., if $a$ and $b$ both belong to $T_{i}$ then a constraint
$ML(a,b)$ is created. 
{\bf CL constraints:} data elements belonging to different categories $T_{i}$ and
$T_{j}$ are encoded as CL constraints, i.e., if $a \in T_{i}$ and $b \in T_{j}$ and
$i \neq j$, then $CL(a,b)$ is created. 

An important point to note is that data elements of the miscellaneous $Z$ category 
are not encoded with any constraints. In the unsupervised learning process these data 
points have the freedom to move to any cluster.

\subsection{Constrained Clustering}
For constrained clustering we use Davidson and Ravi's  CVEQ (Constrained Vector Quantization Error) formulation to extend the  $k$-means objective to capture both ML and CL constraints~\cite{davidsonR05}.

Let $d(x,y): D \times D \rightarrow R$ be the distance function in the context of the application. 
For a clustering problem on data set $D$ with $k$ clusters, let $g: D \rightarrow \{1,\ldots, k\}$ be the mapping from $D$ to cluster labels. Let $C =\{c_{1},\ldots, c_{k}\}$ be a
set of representative cluster centroids and let $h: C \rightarrow C$ be a function such that $h(c)$ is the nearest centeroid to $c$ for $h(c) \neq c$. 
Finally, let $\pi$ represent a cluster set.
We define the {\em error} objective function of assigning $D$ to $C$ as 

\begin{align}
E(C,D)  =  
 & \frac{1}{2}\sum_{j=1}^{k}\left\{\sum_{x_{i} \in \pi_{j}} d^{2}(x_{i},c_{j}) \right. + \\
& \sum_{\substack{x \in \pi_{j} \\
 (x_{i},x_{a}) \in ML \\
 g(x_{i}) \neq g(x_{a})}} d^{2}(c_{g(x_{a})},c_{j}) + \\
& \sum_{\substack{x_{i} \in \pi_{j} \\
 (x_{i},x_{a}) \in CL \\
  g(x_{i}) =  g(x_{a})}}\left. d^{2}(c_{h(g(x_{a}))},c_{j}) \right\}
\end{align}

The intuition behind the design of $E(C,D)$ is that besides the standard distortion
error (the first term), if an ML constraint $(x_{i},x_{a})$  is violated then the cost of 
the violation is equal to the distance between the two centroids that contain the instances
which should have been together. Similarly if a CL constraint $(x_{i},x_{a})$
is violated then the error cost is the distance between the centroid $c$ assigned
to the pair and its nearest nearest centroid $h(c)$.

Based on the error objective $E(C,D)$ a $k$-means style of algorithm can be designed which iterates, until convergence, between an assignment and update rule defined as follows~\cite{davidsonR05}: 


\noindent
{\bf Assignment Rule:} 
\begin{gather}
 \forall x_{i} \notin ML \cup CL: \argmin_{j}d^{2}(x_{i},c_{j}) \label{noconstraint}  \\
 \forall (x,y) \in ML:   \nonumber \\ 
  \argmin_{i,j}\left\{d^{2}(x,c_{i}) + d^{2}(y,c_{j}) + \neg\delta(x,y)\ast d^{2}(c_{i},c_{j})\right\} \label{mlconstraint}  \\
 \forall (x,y) \in CL:  \nonumber \\
  \argmin_{i,j}\left\{d^{2}(x,c_{i}) + d^{2}(y,c_{j}) + \neg\delta(x,y)\ast d^{2}(c_{i},h(c_{i}))\right\}  \label{clconstraint}   
\end{gather}
Here $\delta(x,y)$ is the Kronecker delta function, i.e., $\delta(x,y) = 1$ if $ x = y$ and
$\delta(x,y) = 0$ if $ x \neq y$. Thus if $(x,y)$ is an ML constraint and $x \neq y$,
then $\neg\delta(x,y) = 1$ and if $x$ and $y$ end up belonging to different clusters ($c_{i}$
and $c_{j}$) then an error of $d^{2}(c_{i},c_{j})$ is incurred. Similarly, if 
$(x,y)$ in a CL constraint and $x$ and $y$ end up in the same cluster then
a cost of $d^{2}(c_{i},h(c_{i}))$ is incurred where $h(c_{i})$ is the nearest cluster
centroid to $c_{i}$. Note each of the $\argmin$ operators in the above
assignment rules will output the centroid or pair of centroids.\\[2ex]
\noindent
{\bf Update Rule:} \\
\begin{align}
c_{j} = \frac{\displaystyle
\sum_{x_{i} \in \pi_{j}}\left[x_{i}\right. + \sum_{\substack{(x_{i},y) \in ML, \\
  g(x_{i}) \neq g(y)}}c_{g(y)} +
 \sum_{\substack{(x_{i},y) \in CL, \\
g(x_{i}) = g(y)}}\left. c_{h(g(y))}\right]} 
{\displaystyle
|\pi_{j}| + \sum_{\substack{(x_{i},y) \in ML, \\
  g(x_{i}) \neq g(y)}}1 +
 \sum_{\substack{(x_{i},y) \in CL, \\
 g(x_{i}) = g(y)}}1}
\label{urule}
\end{align}
The update rule of $c_{j}$  computes a modified average of all points that
belong to $\pi_{j}$. The modification captures the number of elements in $\pi_{j}$
which violated the ML and CL constraints.
\subsection{The COD-Means Problem }
In order to formally capture the tendency of humans (and the SL algorithm)) to 
make errors during the labeling process, we introduce a new computational
problem to capture clustering, constraints and outliers.

\begin{problem} (COD-Means)
Given a data set $D$, a distance function $d: D \times D \rightarrow R$, constraint
sets $ML$ and $CL$, parameters 
$k$ and $\ell$ find a set $C =\{c_{1},\ldots,c_{k}\}$ and a set $\cal{L}$ consisting
of $k\times\ell$ points ($\ell$ points per cluster) in order to minimize the error
\end{problem}
\begin{equation}
E(D,C,\cal{L}) = E(D \setminus \cal{L},C)
\end{equation}

\noindent
{\bf Observation:}
The COD-Means problem is NP-hard for $k > 1$ and $\ell \geq 0$. This is clear because
without the outliers the problem is standard clustering with constraints which
is known to be NP-hard in the presence of CL constraints.

\subsection{Algorithm}
We propose a natural extension of the constrained clustering algorithm to
minimize $E(D,C,L)$  shown in Algorithm ~\ref{cod_kmeans}. The algorithm is
similar to a standard clustering algorithm with assignment rules
given in Equation 4 - 6. The key difference
is in Lines 9 - 13, where the points in each cluster $\pi$  are first sorted
based on their distance to the centroid $c_{\pi}$ and the top $\ell$ points
are removed from $\pi$ and inserted into $L(\pi)$. The update rule (Equation 7)
is then applied to the modified $\pi$. 

Note that because of the presence of ML and CL constraints and that they have to be 
processed in pairs, the running time of the algorithm is bounded by $O(|D|^{2}k^{2}I)$ 
where $I$ is the number of iterations. We have omitted the convergence analysis
of the algorithm due to space limitations; it is an extension of ~\cite{chawla2013k}.

\begin{algorithm}
\begin{algorithmic}[1]
\REQUIRE{Data $D$, $ML$ and $CL$ constraints on $D$, $k$ number of clusters, $l$ number of 
outliers per cluster}
\ENSURE{Cluster sets $\Pi$, Outlier sets  $\cal{L}$}
\STATE Initialize with $c_{1},\ldots, c_{k}$ centroids 
\WHILE {(not converged)}
\FORALL { $x \notin ML \cup CL$}
\STATE Use Assignment Rule (Eqn \ref{noconstraint}) 
\ENDFOR
\FORALL { $(x,y) \in ML$ }
\STATE Use Assignment Rule (Eqn \ref{mlconstraint}) 
\ENDFOR
\FORALL {$(x,y) \in CL$}
\STATE Use Assignment Rule (Eqn \ref{clconstraint}) 
\ENDFOR
\FORALL {$ \pi \in \Pi$} 
\STATE Re-order points $x_i$ in $\pi$ such that
\STATE $d(x_{1},c(\pi)) \geq d(x_{2},c(\pi))\ldots, \geq d(x_{|\pi|},c(\pi))$
\STATE $L({\pi}) = \{x_{1},\ldots,x_{\ell}\}$
\STATE $\pi = \pi \setminus L({\pi})$
\STATE Update $c_{\pi}$ using Update Rule (Eqn \ref{urule})
\ENDFOR
\ENDWHILE
\STATE $\Pi = \{\pi_{1},\ldots \pi_k\}$
\STATE $\cal{L}$ $= \{L(\pi_{1}),\ldots,L(\pi_k)\}$
\end{algorithmic}
\caption{The COD-Means Algorithm}
\label{cod_kmeans}
\end{algorithm}

%% file: 04-experiments.tex
\section{Experiments}\label{sec:experiments}
We have designed and executed an extensive set of experiments to validate the
proposed Expert-Machine-Crowd (EMC) framework. In particular we would like
to resolve the following questions:
\begin{enumerate}
\item Are the new clusters identified by the {\tt COD-Means} algorithm genuinely
different and novel compared to the topical clusters (i.e. existing categories) previously defined?

\item What is the nature of outliers (i.e. in our case represent labeling errors) discovered by the {\tt COD-Means} algorithm.
Are they genuine outliers, i.e., they do not (semantically) belong to the clusters.

\item What is the impact of outliers on the quality of clusters generated
from the {\tt COD-Means} algorithm?

\item Once the new clusters are added to the training process and the outliers/wrong labels removed,
does the overall accuracy of the classification process improve?
\end{enumerate} 
\begin{figure*}[htbp]
\centering
\includegraphics[width=0.85\textwidth]{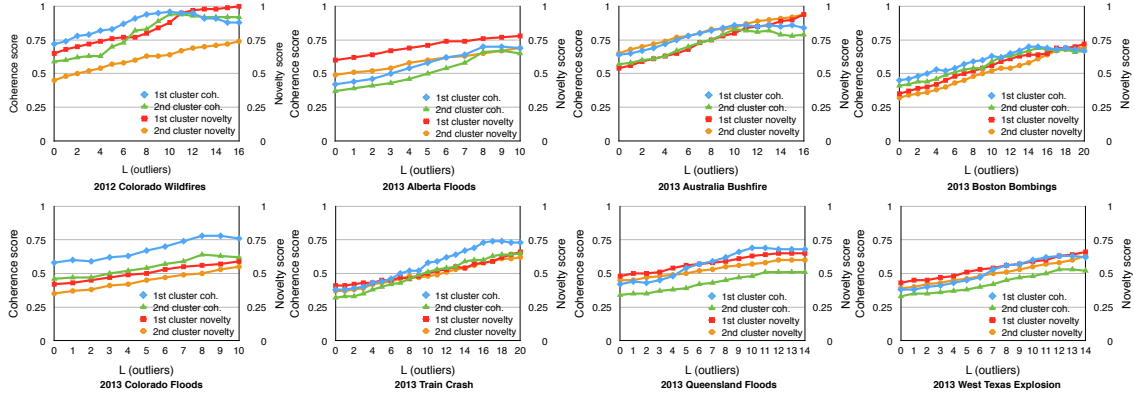}
\caption{Coherence and novelty scores using different values of parameter $\ell$ (outliers) of two largest clusters using simple $k$-means and {\tt COD-means} clustering algorithms.\label{fig:coh_nov_fig}}
\end{figure*}

\subsection{Datasets}\label{data}
We use 8 datasets from~\cite{olteanu2015expect}, which corresponds to English messages posted on Twitter during crises in 2012 and 2013. 
Each crisis in this collection comprises 1,000 tweets annotated using the following categories: 
$T_A$: Affected individuals, $T_B$: Infrastructure and utilities, $T_C$: Donations and volunteering, $T_D$: Caution and advice, $T_E$: Sympathy and emotional support, $Z$: Miscellaneous.

Table \ref{tab:datasets} lists the crises and the prevalence of the topical ($T_i$) and miscellaneous ($Z$) categories. 

\begin{table}[!thbp]
\centering
\scriptsize
\caption{Datasets details: crisis name, year, \% of tweets.
}
\label{tab:datasets}
\begin{tabular}{llll}
\hline
Crisis name              & Year & \multicolumn{1}{c}{\begin{tabular}[c]{@{}l@{}}\% in\\ $T_A..T_E$\end{tabular}} & \multicolumn{1}{c}{\begin{tabular}[c]{@{}l@{}}\% in\\ Z\end{tabular}} \\ \hline
Colorado wildfires (CWF)       & 2012 & 55\%                                                                                          & 45\%                                                                                              \\
Alberta floods (AF)           & 2013 & 82\%                                                                                          & 18\%                                                                                              \\
Australia bushfires (ABF)       & 2013 & 60\%                                                                                          & 40\%                                                                                              \\
Boston bombings (BB)          & 2013 & 59\%                                                                                          & 41\%                                                                                              \\
Colorado floods (CF)          & 2013 & 76\%                                                                                          & 24\%                                                                                              \\
NY train crash (NYTC)           & 2013 & 54\%                                                                                          & 46\%                                                                                              \\
Queensland floods (QF)        & 2013 & 70\%                                                                                          & 30\%                                                                                              \\
West Texas explosion (WTE)     & 2013 & 77\%                                                                                          & 23\% \\   \hline
\end{tabular}
\end{table}

\subsection{Comparison with standard $k$-means (clusters quality)}\label{sec:baseline}

First we compare our proposed algorithm with the baseline algorithm i.e. standard $k$-means. For this purpose, we use standard cluster metrics: cluster cohesiveness and novelty. 

\spara{Cohesiveness.} We measure cohesiveness based on standard intra- and inter-similarity metrics.
For a cluster $C_i$, we denote by $\intra(C_{i})$ its intra-cluster distance, defined as the average distance of elements inside the cluster:
\[
\intra(C_{i}) = \frac{\sum_{a,b \in C_{i}} d(a,b)}{|C_i|^2}~.
\]

For $C_i$, we denote by $\inter(C_{i}, C^c_{i})$ its average inter-cluster distance with respect to elements in the pre-existing categories:
\[
\inter(C_{i},\cup_{i=1}^k T_i) = \frac{\sum_{a \in C_{i}, b \in \cup_{i=1}^k T_i} d(a,b)}{|C_{i}||\cup_{i=1}^k T_i|}~.
\]
We ignore the $Z$ category in this inter-cluster calculation, as it does not represent a specific topic, but corresponds to elements that do not fit in any of the existing topics.
The cohesiveness of a cluster $C_i$ is defined as a combination of the cluster intra-similarity and its inter-similarity with other clusters: $\intra(C_{i}) / \inter(C_{i},C^c_{i})$.
Ideally, a cluster/category should have high cohesiveness, i.e. small intra-cluster distance and large inter-cluster distances.

\spara{Novelty.}
A candidate category must not to be similar to previously existing categories i.e. it must be novel. Similar categories confuse human annotators and reduce the effectiveness of an automatic classifier.
The novelty of a cluster $C_{i}$ is determined as $\maxDist(C_i) - \minDist(C_i)$ where:
\[
\maxDist(C_i) = \max_{a \in C_{i}, b \in \cup_{i=1}^k T_i} d(a,b)
\]
and
\[
\minDist(C_i) = \min_{a \in C_{i}, b \in \cup_{i=1}^k T_i} d(a,b)~.
\]

A high novelty is observed when a cluster is far apart from all the other pre-existing clusters $T_i$.

\spara{Comparison.}
To generate clusters using both approaches ($k$-means and {\tt COD-means}), we generate $k_o+n+1$ clusters, where $k_o$ is the number of pre-existing non-miscellaneous categories (i.e. 5 in our case), $n$ is the number of new categories we aim to generate from the $Z$ category. In this case, we use $n=4$ which is set heuristically, as we observe that slightly larger values (from 5 to 10) do not yield significantly different results and can cause labeling errors (as discussed above), and smaller values tend to yield very general categories. The $+1$ corresponds to the new miscellaneous category $Z'$.
We vary the number of outliers $\ell$ from $\ell=0 \dots m$.

Results generated using $\ell=0$ represent the standard $k$-means algorithm i.e. without outliers detection. We compute cohesiveness and novelty scores for the clusters generated using each value of $\ell$. We pick the ``top'' clusters from the output. We define a ``top'' cluster as one that is either among the $m$ having the largest coherence, the $m$ having the largest novelty, or both, with $m<k$. We use $m=2$ to reduce the number of annotations needed.


Figure~\ref{fig:coh_nov_fig} depicts the results for all the datasets. The proposed approach generates more cohesive and novel clusters by removing outliers. As the value of $\ell$ increases, more tight and coherent clusters are observed. This also helped us determine an optimal value of the $\ell$ parameter for each dataset (i.e. where the high coherence and novelty scores were noticed) to be used in the next experiments. 

\subsection{Data improvements evaluations}
To achieve high classification accuracy, we aim to discover and remove incorrectly categorized items to miscellaneous and non-miscellaneous categories. 
We perform the following two data improvements experiments.

\subsubsection{Labeling errors in non-miscellaneous categories}\label{sec:outlier_of_abc}

As described earlier {\tt COD-Means} discovers local outliers for each newly generated cluster. To determine whether outliers of the non-miscellaneous clusters are semantically
genuine outliers, which we also call labeling errors, we have performed a user study.


To generate clusters and discover outliers, we ran {\tt COD-Means}  using $k=k_o+n+1$ where $k_o$ represents the number of non-miscellaneous categories i.e. $5$ in all of our datasets. 
As in this evaluation we are not interested in generating new clusters from $Z$, we set $n=0$, and as always we keep $+1$, that makes $k=6$. The optimal values for $\ell$ identified in the previous section were used for each dataset. Labeling errors from each non-miscellaneous clusters are obtained. Table~\ref{tab:outliers} shows a few examples of wrongly labeled items identified by {\tt COD-Means}.

\begin{table*}[htpb]
\centering
\scriptsize
\caption{Outliers (labeling errors) identified by the {\tt COD-Means} algorithm in the datasets (user mentions and URLs are masked)}
\label{tab:outliers}
\begin{tabular}{lll}
Dataset   & Actual category      & Outlier                                                                                                                 \\ \hline
2012\_QWF  & Sympathy and support  & \begin{tabular}{@{}c@{}}{\tt What a week it's been! \#highparkfire you better behave today. http://.../} \end{tabular}\\

2013\_AF  & Sympathy and support & {\tt \begin{tabular}{@{}l@{}} What I didn't think about when leaving yesterday electricity shut off...goodbye food in \\ fridge and freezer \#yycflood \end{tabular}}\\

2013\_ABF & Affected individual  & {\tt \begin{tabular}{@{}l@{}} RT @UserMention: Another shot of the cricket at \#Faulconbridge A full pitch view \\ Photographer- @UserMention \#nswfires @UserMention http://.../ \end{tabular}} \\

2013\_CF & Affected individual  & {\tt \#coloradoflood : Stories of grief, generosity http://.../} \\


2013\_NYTC  & Affected individual  & {\tt @NTSB: Alcohol tests on \#MetroNorth train crew all negative after crash} \\


2013\_QF  & Infrastructure and utilities  & {\tt \begin{tabular}{@{}l@{}} RT @UserMention: Our partners w/@UserMention created this map of the West, TX explosion\end{tabular}} \\
\hline
\end{tabular}
\end{table*}

Next, we generate crowdsourcing tasks using the identified outliers. A crowdsourcing task consists of an outlier item, its actual category (i.e. name of a $T_i$), and the category description. Workers were asked to read the category description and choose whether the given item is related to the category, or not. We used the CrowdFlower\footnote{http://crowdflower.com/} crowdsourcing platform. 
At least three different workers were required to finalize a task. For this task the inter-annotator agreement was 78\%.

\begin{figure}[t]
\centering
\includegraphics[width=\columnwidth]{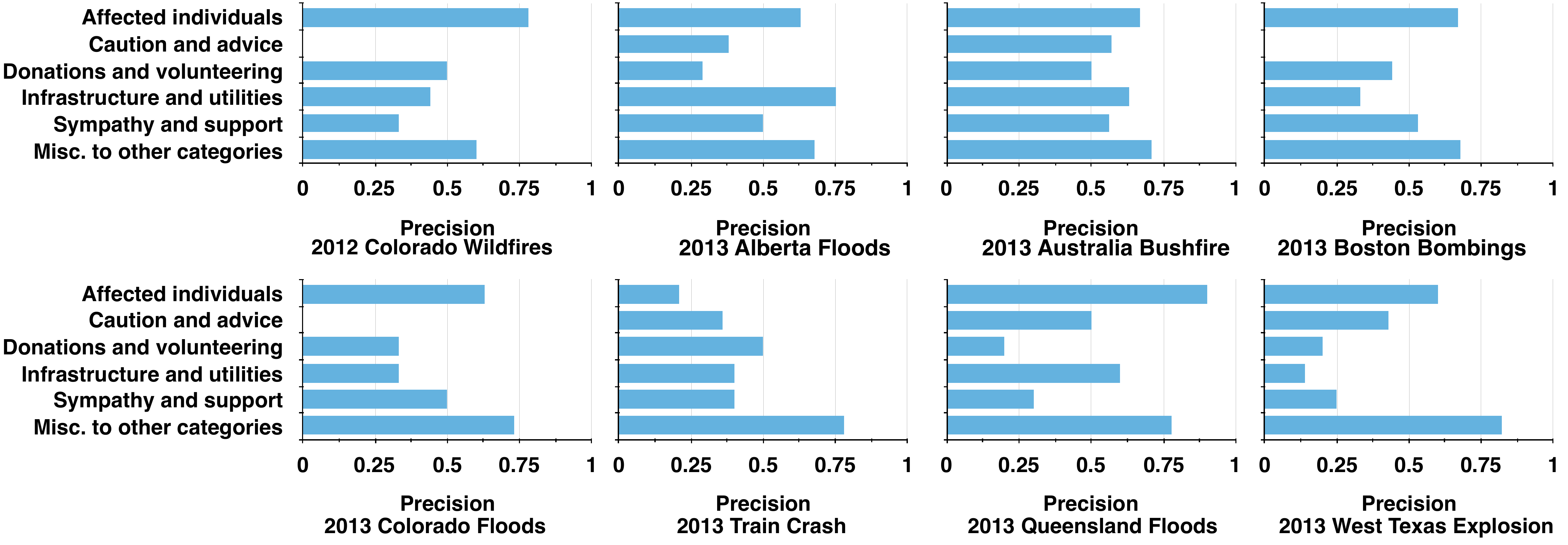}
\caption{Precision scores of outliers identified in non-misc. categories and items moved from misc. to other categories\label{fig:precision_abc_outliers}}
\end{figure}

Figure~\ref{fig:precision_abc_outliers} depicts the results obtained from the crowdsourcing task in terms of precision. The precision $P$ is measured as:
$P = \frac{\mathrm{number\: of\: correctly\: identified\: outliers}}{\mathrm{number\: of\: identified\: o utliers}}$.
From the results (Figure~\ref{fig:precision_abc_outliers} and Table~\ref{tab:outliers}) we can see that the proposed approach can help discover real labeling errors. To achieve a high classification accuracy, wrongly labeled items must be discarded to help classifier achieve higher generalization accuracy~\cite{cao2003modified}, which we test in section~\ref{sec:labeling_errors}.

\subsubsection{Items incorrectly labeled as miscellaneous}\label{sec:z_to_abc}

In this task, we aim to evaluate whether items moved from the miscellaneous category to one of the non-miscellaneous categories are genuinely correct. We ran the {\tt COD-means} algorithm using the same settings specified in section~\ref{sec:outlier_of_abc}. We asked crowd workers to specify whether an item's newly assigned category is correct or not. 
The inter-annotator agreement for this task was 76\%.


Figure~\ref{fig:precision_abc_outliers} shows the results (see precision scores against ``misc. to other category"). The precision $P$ is measured as: $P = \frac{\mathrm{number\: of\: correctly\: assigned\: items}}{\mathrm{number\: of\: assigned\: items}}$ The results clearly show that a large proportion of newly assigned items to the non-miscellaneous categories are indeed correct assignments. Hence, if used as a training set, they can boost classification accuracy, which we empirically prove in the next section.

\subsection{Evaluation in terms of utility as training sets}\label{sec:labeling_errors}

To further validate the results obtained from the data improvements process, we train machine learning classifiers to compare classification accuracy before and after data improvements. 

Two training sets are formed (i.e. original labels and labels after the data improvements process). We remove stop-words, URLs, and user mentions from the items. Stemming is performed using the Lovins stemmer. Uni-grams and bi-grams are used as features and we used the information gain feature selection method to select top 1k features. We use three well-known learning algorithms: SVM (support vector machines), Naive Bayes (NB), and Random forest (RF). We perform the evaluation of the learned model using 10-fold cross validation. 
%
%
Table~\ref{tab:classification-2} shows the classification results in terms of AUC. 
A substantial gain in AUC can be clearly observed in case of the improved training data.

\begin{table}[ht!]
\centering
\caption{Classification accuracies in terms of AUC for all the datasets before and after the data improvements process (Results of the Sympathy and support class are omitted due to space limitations).
}
\tiny
\label{tab:classification-2}
\begin{tabular}{llllllllllll}
\hline
\multicolumn{2}{|l|}{\begin{tabular}[c]{@{}l@{}}Categories $T_i$\end{tabular}} & \multicolumn{2}{l|}{\begin{tabular}[c]{@{}l@{}}Affected\\ individuals\end{tabular}} & \multicolumn{2}{l|}{\begin{tabular}[c]{@{}l@{}}Caution and\\ advice\end{tabular}} & \multicolumn{2}{l|}{\begin{tabular}[c]{@{}l@{}}Donations and\\ volunteering\end{tabular}} & \multicolumn{2}{l|}{\begin{tabular}[c]{@{}l@{}}Infrastructure\\ and utilities\end{tabular}} \\ \hline
\multicolumn{1}{|l|}{Dataset}              & \multicolumn{1}{l|}{Classifier}              & \multicolumn{1}{l|}{Before}              & \multicolumn{1}{l|}{After}              & \multicolumn{1}{l|}{Before}              & \multicolumn{1}{l|}{After}             & \multicolumn{1}{l|}{Before}                  & \multicolumn{1}{l|}{After}                 & \multicolumn{1}{l|}{Before}                   & \multicolumn{1}{l|}{After}                 \\ \hline
                                    & NB                                  & 0.77                                   & \textbf{0.83}                         & 0.80                                   & \textbf{0.92}                        & 0.87                                       & \textbf{0.90}                            & 0.85                                        & \textbf{0.89}                             \\
CWF                                      & RF                                & 0.77                                   & \textbf{0.81}                         & 0.68                                   & \textbf{0.90}                        & 0.83                                       & \textbf{0.85}                            & 0.84                                        & \textbf{0.86}                            \\
                                               & SVM                                          & 0.76                                   & \textbf{0.79}                         & 0.74                                   & \textbf{0.91}                        & 0.81                                       & \textbf{0.87}                            & 0.85                                        & \textbf{0.87}                             \\ \hline
                                               & NB                                  & 0.84                                   & \textbf{0.87}                         & 0.72                                   & \textbf{0.74}                        & 0.74                                       & \textbf{0.92}                            & 0.57                                        & \textbf{0.86}                             \\
AF                                       & RF                                & 0.68                                   & \textbf{0.83}                         & 0.58                                   & \textbf{0.71}                        & 0.80                                       & \textbf{0.89}                            & 0.70                                        & \textbf{0.84}                             \\
                                               & SVM                                          & 0.81                                   & \textbf{0.84}                         & 0.67                                   & \textbf{0.72}                        & \textbf{0.88}                              & 0.87                                     & 0.79                                        & \textbf{0.82}                            \\ \hline
                                               & NB                                  & 0.74                                   & \textbf{0.81}                         & 0.76                                   & \textbf{0.82}                        & 0.88                                       & \textbf{0.91}                             & 0.78                                        & \textbf{0.83}                              \\
ABF                                      & RF                                & 0.66                                   & \textbf{0.73}                         & 0.77                                   & \textbf{0.81}                        & 0.83                                       & \textbf{0.91}                            & 0.72                                         & \textbf{0.81}                             \\ 
                                               & SVM                                          & 0.69                                   & \textbf{0.78}                         & 0.76                                   & \textbf{0.79}                        & 0.83                                       & \textbf{0.90}                            & 0.74                                        & \textbf{0.81}                            \\ \hline
                                               & NB                                  & 0.80                                   & \textbf{0.84}                         & 0.66                                   & \textbf{0.93}                        & 0.88                                       & \textbf{0.91}                            & 0.68                                        & \textbf{0.97}                             \\
BB                                       & RF                                & 0.75                                   & \textbf{0.87}                         & 0.56                                   & \textbf{0.94}                        & 0.75                                       & \textbf{0.87}                            & 0.48                                        & \textbf{0.96}                             \\ 
                                               & SVM                                          & 0.73                                   & \textbf{0.81}                         & 0.55                                   & \textbf{0.91}                        & 0.79                                       & \textbf{0.92}                            & 0.59                                        & \textbf{0.97}                             \\ \hline
                                               & NB                                  & 0.91                                   & \textbf{0.94}                         & 0.78                                   & \textbf{0.85}                        & 0.91                                       & \textbf{0.92}                            & 0.84                                        & \textbf{0.88}                             \\
CF                                       & RF                                & 0.89                                   & \textbf{0.92}                         & 0.72                                   & \textbf{0.82}                        & 0.90                                       & \textbf{0.92}                            & 0.80                                        & \textbf{0.84}                             \\ 
                                               & SVM                                          & 0.87                                   & \textbf{0.90}                         & 0.72                                   & \textbf{0.80}                        & 0.90                                       & 0.90                                     & 0.81                                        & \textbf{0.85}                             \\ \hline
                                               & NB                                  & 0.93                                   & 0.93                                  & \textbf{0.72}                          & 0.68                                 & 0.71                                       & 0.71                                     & 0.91                                        & \textbf{0.92}                             \\
NYTC                                     & RF                                & 0.93                                   & \textbf{0.94}                         & 0.58                                   & \textbf{0.60}                        & 0.49                                       & 0.49                                     & 0.83                                        & \textbf{0.89}                            \\ 
                                               & SVM                                          & \textbf{0.93}                          & 0.92                                  & 0.53                                   & \textbf{0.54}                        & 0.48                                       & 0.48                                     & 0.74                                        & \textbf{0.88}                             \\ \hline
                                               & NB                                  & 0.73                                   & \textbf{0.87}                         & 0.77                                   & \textbf{0.79}                        & 0.85                                       & \textbf{0.93}                            & 0.76                                        & \textbf{0.88}                             \\
QF                                       & RF                                & 0.69                                   & \textbf{0.85}                         & 0.69                                   & \textbf{0.73}                        & 0.75                                       & \textbf{0.93}                            & 0.77                                        & \textbf{0.87}                               \\ 
                                               & SVM                                          & 0.69                                   & \textbf{0.86}                         & 0.69                                   & \textbf{0.72}                        & 0.85                                       & \textbf{0.92}                            & 0.75                                        & \textbf{0.87}                           \\ \hline
                                               & NB                                  & 0.89                                   & \textbf{0.93}                         & 0.80                                   & \textbf{0.95}                        & 0.95                                       & \textbf{0.96}                            & 0.82                                        & \textbf{0.93}                              \\
WTE                                      & RF                                & 0.85                                   & \textbf{0.90}                         & 0.61                                   & \textbf{0.91}                        & 0.82                                       & \textbf{0.86}                            & 0.69                                         & \textbf{0.91}                              \\ 
                                               & SVM                                          & 0.80                                   & \textbf{0.86}                         & 0.54                                   & \textbf{0.91}                        & 0.85                                       & \textbf{0.87}                            & 0.74                                        & \textbf{0.92}                             
\\ \hline 

\end{tabular}
\end{table}



%% file: 05-relatedwork.tex

\section{Related Work}\label{sec:relatedwork}

In general cluster analysis methods attempt to form disjoint groups of unlabeled data items such that items in same group are similar while items in different groups are dissimilar. For instance, one such famous clustering method is $k$-means~\cite{hartigan1979algorithm}. However, in a semi-supervised clustering approach, the algorithm uses both labeled and unlabeled data. In this particular case, the labeled data items are used as background knowledge during the clusters generation process. 

Many works described the use of constraints in different ways, e.g. some use constraints at group-level (i.e. on group of items) and other use at item-level (i.e. between two items) as a way to provide background knowledge~\cite{xing2002distance,basu2004active,wagstaff2001constrained}. For instance, in~\cite{wagstaff2001constrained} $k$-means algorithm is extended to use instance-level constraints. However, our method, in addition to the instance-level constraints, discovers and removes outliers during the cluster generation process by which more compact clusters can be obtained.

While it is well-known that outlier detection in a training set can help improve the accuracy of a classifier built using that training set (e.g.~\cite{cao2003modified}), previous methods do not take into consideration the existing categories. Our method unifies constrained clustering and outlier detection by formulating a novel optimization problem and algorithm {\tt COD-Means}.


%% file: 06-conclusions.tex

\section{Conclusions}\label{sec:conclusions}
%

As supervised learning systems can not be used to identify novel concepts, for this purpose, we employ unsupervised learning techniques. We presented a novel clustering algorithm {\tt COD-Means}, which uses human and machine categorized items as background knowledge to form constraints and detects novel categories. The proposed algorithm, not only help detect novel categories, but also seamlessly discover outliers from each cluster by which categorization errors are fixed. Extensive experiments using real datasets demonstrate that our approach is effective and efficient.